\documentclass{article} 
\usepackage{original,times}

\usepackage{amsmath,amsfonts,bm}

\def\eqref#1{equation~\ref{#1}}

\def\1{\bm{1}}

\DeclareMathAlphabet{\mathsfit}{\encodingdefault}{\sfdefault}{m}{sl}
\SetMathAlphabet{\mathsfit}{bold}{\encodingdefault}{\sfdefault}{bx}{n}

\DeclareMathOperator*{\argmax}{arg\,max}

\usepackage{amsmath,amsfonts,bm}

\def\eqref#1{equation~\ref{#1}}

\def\1{\bm{1}}

\DeclareMathAlphabet{\mathsfit}{\encodingdefault}{\sfdefault}{m}{sl}
\SetMathAlphabet{\mathsfit}{bold}{\encodingdefault}{\sfdefault}{bx}{n}

\usepackage{hyperref}
\usepackage{url}
\usepackage[many]{tcolorbox}
\usepackage{xcolor}
\usepackage{tikz}
\usepackage{booktabs}
\usepackage{algorithm}
\usepackage{wrapfig}
\usepackage{float} 
\usepackage{capt-of}
\usepackage{amssymb}
\usepackage{algpseudocode} 
\usepackage{graphicx}
\usepackage{caption}

\newcommand{\roundredbox}[3][red!5]{%
    \begin{center}
    \tikz \node[draw=#2, fill=#1, rounded corners=1mm, inner sep=6pt, text width=0.9\linewidth] {#3};
    \end{center}
}

\newcommand{\roundbluebox}[3][blue!5]{%
    \begin{center}
    \tikz \node[draw=#2, fill=#1, rounded corners=1mm, inner sep=6pt, text width=0.9\linewidth] {#3};
    \end{center}
}

\title{
    PriM: Principle-Inspired Material Discovery through Multi-Agent Collaboration
}

\iclrfinalcopy

\author{
Zheyuan Lai\textsuperscript{*,1,2} \\
\href{mailto:zheyuan_lai@u.nus.edu}{\texttt{zheyuan\_lai@u.nus.edu}}
\And
Yingming Pu\textsuperscript{*,\dag,1,3} \\
\href{mailto:puyingming@westlake.edu.cn}{\texttt{puyingming@westlake.edu.cn}}
\AND
\textsuperscript{1}AMAIR Lab \\
\textsuperscript{2}National University of Singapore \\
\textsuperscript{3}Westlake University and Zhejiang University
}

\begin{document}

\maketitle

\begingroup
\renewcommand\thefootnote{}
\footnotetext{\textsuperscript{*}Equal contribution}
\footnotetext{\textsuperscript{\dag}Corresponding author}
\endgroup

\begin{abstract}
Complex chemical space and limited knowledge scope with biases holds immense challenge for human scientists, yet in automated materials discovery. Existing intelligent methods relies more on numerical computation, leading to inefficient exploration and results with hard-interpretability. To bridge this gap, we introduce a principles-guided material discovery system powered by language inferential multi-agent system (MAS), namely \texttt{PriM}. Our framework integrates automated hypothesis generation with experimental validation in a roundtable system of MAS, enabling systematic exploration while maintaining scientific rigor. Based on our framework, the case study of nano helix demonstrates higher materials exploration rate and property value while providing transparent reasoning pathways. This approach develops an automated-and-transparent paradigm for material discovery, with broad implications for rational design of functional materials. Code is publicly available at our \href{https://github.com/amair-lab/PriM}{GitHub}.
\end{abstract}

\section{Introduction}
The discovery of materials with targeted-property stands as a cornerstone of scientific progress. Yet, the complexity of material space of structures and properties, poses a formidable challenge to traditional discovery paradigms~\citep{Xue2016AcceleratedSF, Wang2022DataDrivenMI}.
Human-driven approaches, while foundational, are inherently constrained by cognitive biases, fragmented knowledge domains, and the practical impossibility of exhaustively probing this multidimensional space~\citep{Brunin2019TransparentCM}.

Contemporary data-driven strategies such as active learning (AL) and reinforcement learning (RL), though powerful, neglect the integration of foundational scientific principles and reduces exploration to a black-box process, prioritizing statistical correlations over mechanism understanding~\citep{Chitturi2023TargetedMD, Lookman2019ActiveLI, Fare2024MANDRELMR, Beeler2024ChemGymRLAC, Kim2023MaterialsDW}.
For instance, while AL iteratively refines search spaces using acquisition functions, it often operates as a reactive framework, lacking proactive hypothesis generation.
Similarly, the reward-driven exploration in RL may optimize toward narrow objectives without contextual alignment with broader material design principles.
This results in inefficiencies, such as redundant sampling of unproductive regions, and solutions that lack interpretability, limiting their utility in guiding actionable scientific insights.
Recent techniques such as tools augmented large language models (LLMs), so-called agents, demonstrate remarkable capability in hypothesis generation and scientific reasoning~\citep{Zhou2024HypothesisGW, Yang2023LargeLM, Luo2025LLM4SRAS, Yang2024MOOSEChemLL, Zhang2024ScientificLL}. However, they are limited to explore the materials space with continuous experiments as feedback in scientific discovery to reach a goal, primarily due the awareness of scientific mechanisms and the design of leveraging scientific principles~\citep{Cohrs2024LargeLM, Kumbhar2025HypothesisGF, Baek2024ResearchAgentIR, Su2024TwoHA}.
These limitations underscore a critical gap:

\vspace{-0.2cm}
\roundredbox{red}{\textbf{Critical gap:} The need for an automated discovery framework that \\ (1) maintains \textbf{scientific rigor} through principle-guided exploration, \\ (2) ensures \textbf{experimental efficiency} via coordinated multi-agent reasoning, while \\ (3) preserving \textbf{mechanistic transparency} for human-interpretable insights.}
\vspace{-0.2cm}

This three-fold challenge necessitates a fundamental rethinking of how automated systems can embed scientific principles while maintaining operational efficiency and interpretability in materials discovery.

\begin{figure}[t!]
    \centering
    \includegraphics[width=1.0\textwidth]{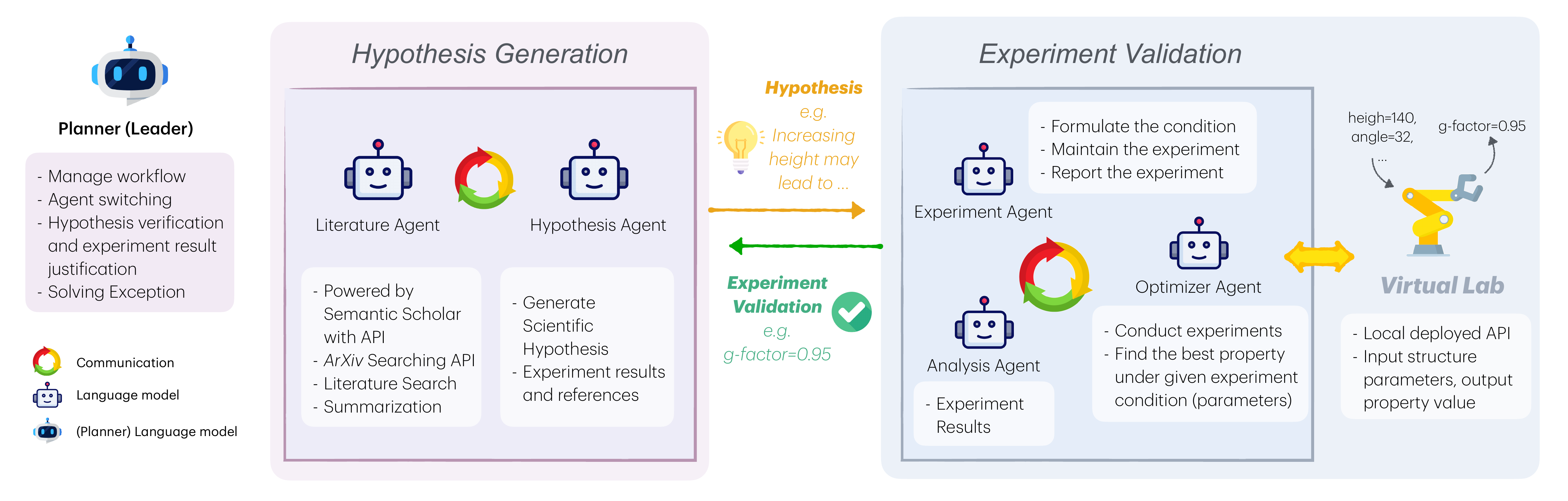}
    \vspace{-0.3cm}
    \caption{
        Overview of \texttt{PriM}.  Two phases: hypothesis generation and experimental validation. The Planner initiates a hypothesis loop involving a Literature Agent (gathering prior knowledge) and Hypothesis Agent (formulating testable hypotheses). Validated hypotheses then undergo experimental testing: the Experiment Agent designs conditions, while the Optimizer Agent employs method, i.e., Monte Carlo Tree Search (MCTS), to optimize outcomes. The process iterates through reasoning and roundtable discussions to refine experiments. Crucially, hypotheses are grounded in physicochemical principles, balancing exploration and exploitation to drive discovery.
    }
    \label{fig:schema}
\end{figure}

To address this, we propose a paradigm shift toward a scientific principles-guided materials discovery framework, powered by a language-inferential MAS, namely \texttt{PriM}. Unlike conventional numerical approaches, our workflow embeds domain knowledge and physicochemical principles into the exploration process, enabling agents to generate hypotheses with searched literatures or conducted experiments, reason through natural language context, and collaboratively refine strategies under iterative roundtable. With curly designed prompt engineering, this system integrates two phases, i.e., hypothesis generation and experimental validation, creating a closed-loop cycle that balances exploration with rigorous scientific reasoning. This design gives hypothesis-oriented working guidance for agent collaboration. By leveraging language models for reasoning, \texttt{PriM} provides explicit, human-readable decision pathways, bridging the gap between data-driven optimization and subject-related knowledge. In summary, our contributions are as follows:

\vspace{-0.2cm}
\begin{enumerate}
    \item A novel scientific discovery framework, \texttt{PriM}, that synergies principle-guided exploration with multi-agent systems for automated materials discovery.
    \item Empirical validation through nanohelix material discovery demonstrates 56.3\% improvement in property optimization compared to Vanilla Agent and 9.1\% improvement over Vanilla MAS, highlighting the effectiveness of principle-guided exploration.\@.
    \item Comprehensive experiments establish advantages of \texttt{PriM} in both computational efficiency and scientific interpretability over existing methods.
\end{enumerate}

\section{Related Work}
\subsection{Language models for materials discovery}
Recent language model (LM) advances have enhanced hypothesis generation in materials science through knowledge distillation from literature \citep{Yang2024MOOSEChemLL, Zhou2024HypothesisGW, pu2024leveraging}. While retrieval-augmented LMs like DARWIN and HoneyComb \citep{Xie2023DARWINSD, Zhang2024HoneyCombAF} improve domain-specific reasoning, they struggle to integrate prior principles or experimental constraints, risking hallucinated proposals \citep{Bran2023AugmentingLL}. Hybrid frameworks address this by coupling LMs with computational tools \citep{Beeler2024ChemGymRLAC, Fare2024MANDRELMR}, yet remain limited to single-agent architectures that lack systematic exploration of complex chemical spaces \citep{Luo2025LLM4SRAS}. Surveys highlight persistent gaps in interpretability \citep{Zhang2024ACS, Han2024FromGT}, underscoring the need for principled guidance beyond tool chaining \citep{Ramos2024ARO}.

\subsection{Multi-agent system for scientific discovery}
Multi-agent systems (MAS) show promise in decomposing discovery tasks \citep{Lu2024TheAS, Su2024TwoHA, wang2024survey}, but existing implementations prioritize either parallel experimentation \citep{Li2020TowardM} or black-box optimization \citep{Lookman2019ActiveLI}, lacking transparent reasoning pathways. Physics-aware MAS architectures \citep{Ghafarollahi2024ProtAgentsPD, Ghafarollahi2024SciAgentsAS, ghafarollahi2024atomagents} incorporate domain knowledge but rely on rigid workflows that limit adaptive hypothesis generation. Modular frameworks like HoneyComb \citep{Zhang2024HoneyCombAF} enable tool integration but face coordination inefficiencies in cross-agent validation. While recent works advocate LLM-enhanced agents \citep{Baek2024ResearchAgentIR, Kumbhar2025HypothesisGF}, none synergize language-guided hypothesis space construction with physics-based verification—a critical gap our work addresses.

Prior efforts either focus on data-driven exploration \citep{Kim2023MaterialsDW, schmidgall2025agent, Gomes2019CRYSTALAM} or human-AI collaboration \citep{Chitturi2023TargetedMD, Ni2024MatPilotAL} without unifying interpretable reasoning and automated validation. Our principles-guided MAS uniquely combines (1) language-inferential hypothesis generation constrained by physicochemical principles, (2) dynamic agent societies for parallelized exploration-exploitation, and (3) closed-loop validation with explainable decision trails—advancing beyond single-agent RL \citep{Fare2024MANDRELMR} and static MAS designs \citep{Su2024TwoHA, Lu2024TheAS}. This design directly resolves the efficiency-interpretability trade-off in \citep{Xue2016AcceleratedSF} while outperforming Bayesian optimization \citep{Chitturi2023TargetedMD} in multi-fidelity settings.

%
%
%
\section{Methodology}
\subsection{System Overview}
Our framework, \texttt{PriM}, employs a LLM-based MAS to explore the materials structural-property space. As depicted in Figure~\ref{fig:schema}, the framework comprises two primary components: hypothesis generation and experimental validation, coordinated by a central Planner agent that orchestrates the workflow and validates experimental outcomes.

\subsection{Architecture}
We formally define \texttt{PriM} as $\mathcal{F} = \{\mathcal{P}, \mathcal{H}, \mathcal{E}, \mathcal{S}\}$, where $\mathcal{P}$ denotes the central Planner agent, $\mathcal{H}$ represents the hypothesis generation phase, $\mathcal{E}$ encompasses the experimental validation phase, and $\mathcal{S}$ defines the state space.

\noindent \textbf{Hypothesis generation.} The hypothesis generation phase $\mathcal{H} = \{L, H\}$ incorporates two essential components: the Literature Agent $L: \mathcal{K} \rightarrow \mathcal{I}$, which maps the knowledge space $\mathcal{K}$ to insights $\mathcal{I}$, and the Hypothesis Agent $H: \mathcal{I} \rightarrow \mathcal{T}$, which generates testable propositions $\mathcal{T}$. This phase starts with the Literature Agent synthesizes insights from existing literature based on the predefined research goal and constraints. Hypothesis Agent here is to formulate testable propositions grounded in physicochemical principles. These LLM-based agents rely on carefully designed prompt engineering and iterative checking mechanism to ensure that hypotheses generated are both innovative and aligned with established scientific knowledge.

\noindent \textbf{Experimental validation.} The experimental validation phase $\mathcal{E} = \{E, V, O, A\}$ integrates four key elements: the Experiment Agent $E: \mathcal{T} \times \mathcal{X} \rightarrow \mathcal{D}$, Virtual Laboratory $V: \mathcal{X} \rightarrow \mathcal{Y}$, Optimizer Agent $O: \mathcal{X} \times \mathcal{Y} \rightarrow \mathcal{X}^*$, and Analysis Agent $A: \mathcal{D} \rightarrow \mathcal{R}$. Here, $\mathcal{X}$ represents the parameter space, $\mathcal{Y}$ denotes the property space, and $\mathcal{R}$ encompasses the analysis reports. Powered by the Experiment Agent, the experimental variables and their initial values are carefully designed based on the hypothesis generated in the first phase. The Virtual Laboratory is used to conduct experiments by given structure parameter and yield the property value. The experiment process is supported by the Optimizer Agent, which is to find the optimal parameters with the best property value. The Optimizer Agent collaborates closely with the Analysis Agent, which interprets the experiment results, identifies the statistical patterns discovered, and generates a research experiment report, which is used for the next iteration to refine hypothesis and experiment.

The iterative process follows the formulation $\mathcal{S}_{t+1} = \mathcal{P}(\mathcal{S}_t, \mathcal{R}_t)$, with each iteration step $t$ consisting of hypothesis generation $\mathcal{T}_t = H(L(\mathcal{K}_t))$, experimentation $\mathcal{D}_t = E(\mathcal{T}_t, \mathcal{X}_t)$, optimization $\mathcal{X}^*_t = O(\mathcal{X}_t, V(\mathcal{X}_t))$, and analysis $\mathcal{R}_t = A(\mathcal{D}_t)$. This mathematical framework ensures systematic scientific exploration while maintaining interpretability through the integration of LLM-based reasoning capabilities.

The framework implements a cyclical workflow wherein hypotheses undergo systematic experimental validation, with results informing subsequent hypothesis generation. This iterative approach facilitates continuous refinement of scientific understanding while adhering to rigorous experimental validation protocols. Inter-agent communication is facilitated through an integrated network of language models, with the Planner agent ensuring coherent system-wide coordination. This architectural design enables \texttt{PriM} to effectively navigate the complex landscape of structure-property relationships and physicochemical principles, synthesizing theoretical insights with empirical validation.

\subsection{Principle-Guided Reasoning}
Traditional materials discovery methods, including RL and MCTS, rely on exhaustive numerical exploration of vast chemical spaces \citep{Xue2016AcceleratedSF, Kim2023MaterialsDW}, suffering from interpretability deficits and inefficiency in constrained design scenarios.

\texttt{PriM} addresses these limitations through a triad of principle-driven mechanisms: (1) The Literature Agent retrieves and distills physicochemical principles from experimental studies and theoretical frameworks; (2) Hypothesis generation is constrained via chain-of-principles prompting \citep{Wei2022ChainOT}, enforcing explicit adherence to symmetry rules, thermodynamic feasibility, and synthesis compatibility through constraint set $\mathcal{C}$; (3) The Analysis Agent validates outcomes through mapping experimental observations to mechanistic models that expose causal relationships. This framework achieves a principles-guided exploration, rather than exploration with only experiment results.

\subsection{Automated Validation Pipeline}
\texttt{PriM} transforms the traditionally iterative validation process through a closed-loop integration of virtual experimentation and strategic optimization. The Experiment Agent operates a physics-informed virtual lab, leveraging surrogate models trained on multi-fidelity datasets \citep{Huang2023ApplicationOM} to predict material properties with experimental accuracy. Concurrently, the Optimizer Agent navigates the constrained parameter space via MCTS \citep{browne2012survey}, prioritizing high-potential candidates identified by the Hypothesis Agent while dynamically pruning suboptimal branches.

\roundbluebox{blue}{\noindent \textbf{$\diamond$ Take-Away:} To identify optimal experimental conditions, our \texttt{PriM} framework employs physicochemical principles for initial hypothesis generation, followed by an efficient numerical optimization process that converges within a small number of rounds.}

%
%
%
\section{Experiment Settings}\label{sec:experiment_settings}
To empirically assess the effectiveness of \texttt{PriM} compared to single experimental agent and basic MAS, we perform experiments using nano helix materials structural-property as a case study. A Virtual Lab is constructed with a trained model that predicts property values based on structural parameters, with further details provided in Appendix \ref{sec:virtual_lab}. The dataset for training this model is from a project in cooperation with a certain institution. The project is currently in the process of submission. Due to the cooperation agreement, the dataset is not yet public. This Virtual Lab serves as a platform for structure-property-driven materials discovery. We compare \texttt{PriM} with a baseline MAS approach under simple-and-workable settings to evaluate its relative efficiency.

We set the following research goal and constraints, which will then be summarized and formalized by UserProxy Agent:

\noindent\textbf{Research Goal (task description):} \textit{Find the structural parameters corresponding to the strongest chirality (g-factor characteristics) in the nanohelix material system.}

\noindent\textbf{Research Constraints:} \textit{Explicitly show the underlying physicochemical principles regarding the structure and property relationships.}

For the Literature Agent, we use LLM to generate the query words, which are then processed through the Semantic Scholar API to retrieve at most 4 relevant publications. The searched literature will be passed to LLM to summarize literature insights. The Hypothesis Agent is carefully designed by prompts, which guides the LLM to generate hypotheses based on research goal and constraints, literature insights, past experiment results, and domain knowledge about the subject.

The Experiment Agent is based on the Virtual Lab, which performs experiments on parameters related to the hypothesis. The Optimizer Agent is implemented based on MCTS, which searches within the pre-defined parameter space and aims to discover the optimal g-factor with the given condition. The Analysis Agent analyzes the experiment results using the data analysis tools, with further details provided in Appendix \ref{sec:experiment_data_analysis}. It will then provide a research experiment report, which in the next iteration will instruct the Hypothesis Agent to revise the hypothesis for improvement.

\noindent \textbf{Agents.} The agents are set to interact based on a LLM-based Planner, which uses latest chatting histories to plan the MAS. We use GPT-4o as the language model for all agents. The prompt engineering for each agent and the Planner can be found at Appendix~\ref{sec:prompt_engineering}.

\noindent \textbf{Baselines.} To evaluate the effectiveness of \texttt{PriM}, we compare its performance against two baseline approaches, with implementations detailed in Appendix~\ref{sec:baseline}:
(1) \textbf{Vanilla Agent}: focused solely on suggesting and conducting experiments with MCTS as parameter searching to find the best g-factor, which serves as an experimental baseline.
(2) \textbf{Vanilla MAS}: a multi-agent system similar to \texttt{PriM} but without the Hypothesis Agent, serving as a general multi-agent framework.
All baselines and \texttt{PriM} conduct MCTS for 100 iterations for a fair comparison.

\noindent \textbf{Evaluation metric.} All experiments are conducted based on nanohelix optimization with Virtual Lab. The key performance metrics include the final material property value (g-factor) $\mu$ and the number of experimental steps required to reach it. Additionally, we define the exploration rate $\epsilon$ to quantify the agent's exploration of experimental conditions by calculating the average pairwise distance between all experiment conditions. The metric is formulated by

\begin{equation}
    \epsilon = \frac{1}{N(N-1)}\sum_{i\neq j} \|\mathbf{x_i} - \mathbf{x_j}\|
\end{equation}

where $N$ is the total number of experiments, \textbf{x}$_i$ and \textbf{x}$_j$ are the experimental conditions (i.e., structural parameters), the sum is calculated over all pairs where $i \neq j$.

%
%
%
\section{Results and Analysis}\label{sec:res_ana}

\subsection{Performance Comparison}

We compare our method \texttt{PriM} with baselines across 5 independent runs with random initializations. Our evaluation focuses on exploration rate and convergence iteration count. Traditional optimization methods, such as Bayes Optimization~\citep{Frazier2018ATO} (BO), Deep Q-Net~\citep{Mnih2015HumanlevelCT} (DQN) serve as traditional baseline comparison, though our primary interest lies in comparing the broader search dynamics rather than strictly numerical performance. This approach enables us to assess how principle-guided exploration compares to conventional search strategies in materials discovery contexts.

\roundbluebox{blue}{\noindent \textbf{$\diamond$ Take-Away:} \texttt{PriM} achieves near-optimal material properties while maintaining scientific rationality, unlike traditional optimization methods. Its lower exploration rate compared to Vanilla MAS demonstrates how principle-guided approaches enable more efficient and targeted parameter space traversal, balancing performance with mechanistic understanding.}

\begin{table}[t!]
\small
\centering
\caption{Comparison of \texttt{PriM} with Baselines Methods (mean $\pm$ std)}\label{tab:result}
\begin{tabular}{@{}lccccc@{}}
\toprule
\textbf{Method} &
\textbf{\begin{tabular}[c]{@{}c@{}}Rationality\end{tabular}} &
\textbf{\begin{tabular}[c]{@{}c@{}}Optimal Value ($\mu$)\end{tabular}} &
\textbf{\begin{tabular}[c]{@{}c@{}}Exploration Rate ($\epsilon$)\end{tabular}} &
\textbf{\begin{tabular}[c]{@{}c@{}}Iteration\end{tabular}} & \\
\midrule
BO                    &  N/A                & 1.081 {\small ($\pm$ 0.065)} & 467.35 ($\pm$ 23.52) & 14.29 ($\pm$ 2.34) \\
DQN                   &  N/A                & 1.050 ($\pm$ 0.021) & 6.75 ($\pm$ 0.30)    & 20.00 ($\pm$ 0.00)    \\
Vanilla Agent         &  Naive Logics       & 0.644 ($\pm$ 0.054) & 24.47 ($\pm$ 7.34)   & 9.20 ($\pm$ 2.56)  \\
Vanilla MAS           &  Naive Logics       & 0.923 ($\pm$ 0.170) & 264.65 ($\pm$ 22.42) & 65.40 ($\pm$ 18.91) \\
\texttt{PriM} (Ours)  &  Principles         & 1.007 ($\pm$ 0.103) & 49.68 ($\pm$ 10.07)  & 85.50 ($\pm$ 8.58) \\
\bottomrule
\end{tabular}
\end{table}

\begin{wrapfigure}{r}{0.5\textwidth} 
    \centering
    \vspace{-0.5em}
    \includegraphics[width=0.49\textwidth]{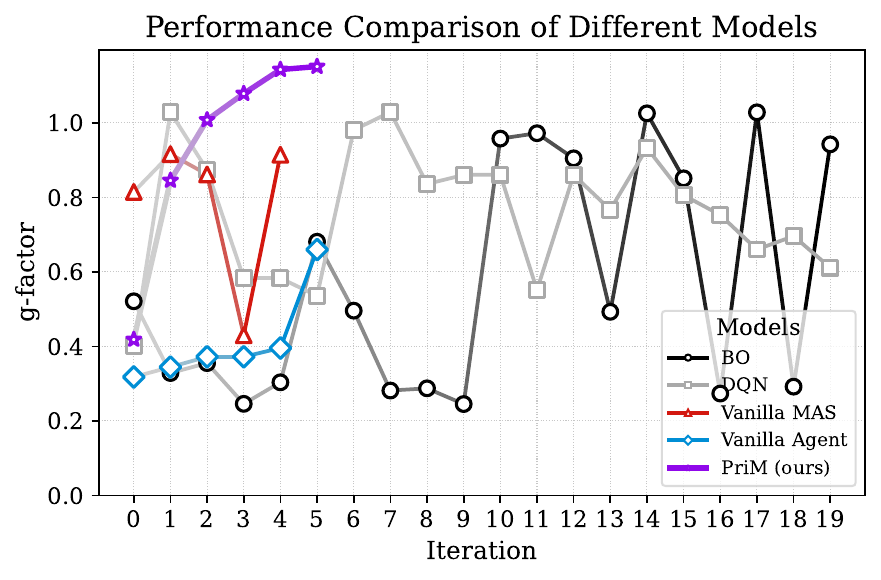}
    \caption{Comparison of nanohelices discovery progress with g-factor optimization. \texttt{PriM} achieves a high g-factor value in significantly fewer steps compared to baseline methods, highlighting the benefits of its physicochemical grounding.}
    \label{fig:comparison}
    \vspace{-0.5em}
\end{wrapfigure}

From Table~\ref{tab:result}, \texttt{PriM} achieves a near-optimal property value (1.007) while significantly outperforming Vanilla Agent, which shows a 36.0\% lower optimal value (0.644). Although traditional optimization methods like BO and DQN reach slightly higher optimal values, they lack scientific rationality in their approach. Notably, while \textit{Vanilla MAS} requires fewer iterations than \texttt{PriM} (65.40 vs. 85.50), it demonstrates a substantially higher exploration rate (264.65 vs. 49.68), indicating inefficient parameter space traversal. This suggests that \textit{Vanilla MAS}, despite exploring more extensively, fails to leverage scientific principles for targeted optimization. As shown in Figure~\ref{fig:comparison}, \texttt{PriM}'s principle-guided approach enables more systematic and efficient exploration, requiring significantly fewer total samples to achieve comparable performance to traditional methods while maintaining interpretable scientific reasoning throughout the discovery process.

These results validate our framework’s design and emphasize the necessity of embedding domain-specific principles into the hypothesis-generation phase. By structuring exploration through scientific reasoning, \texttt{PriM} balances exploration and exploitation more effectively than conventional multi-agent or single-agent approaches, leading to higher optimal material properties and improved interpretability in scientific discovery.

\subsection{Nano Helix Material Discovery}

This case study presents the iterative hypothesis-validation process in nano-helix material discovery, demonstrating how \texttt{PriM} systematically refines hypotheses to achieve an optimized g-factor through principle-guided reasoning. Unlike conventional heuristic-based approaches, \texttt{PriM} follows an iterative cycle where hypotheses are generated, tested, and refined based on both literature insights and experimental feedback, ensuring physicochemically-grounded exploration.

\begin{wrapfigure}{r}{0.5\textwidth} 
    \centering
    \vspace{-0.5em}
    \includegraphics[width=0.49\textwidth]{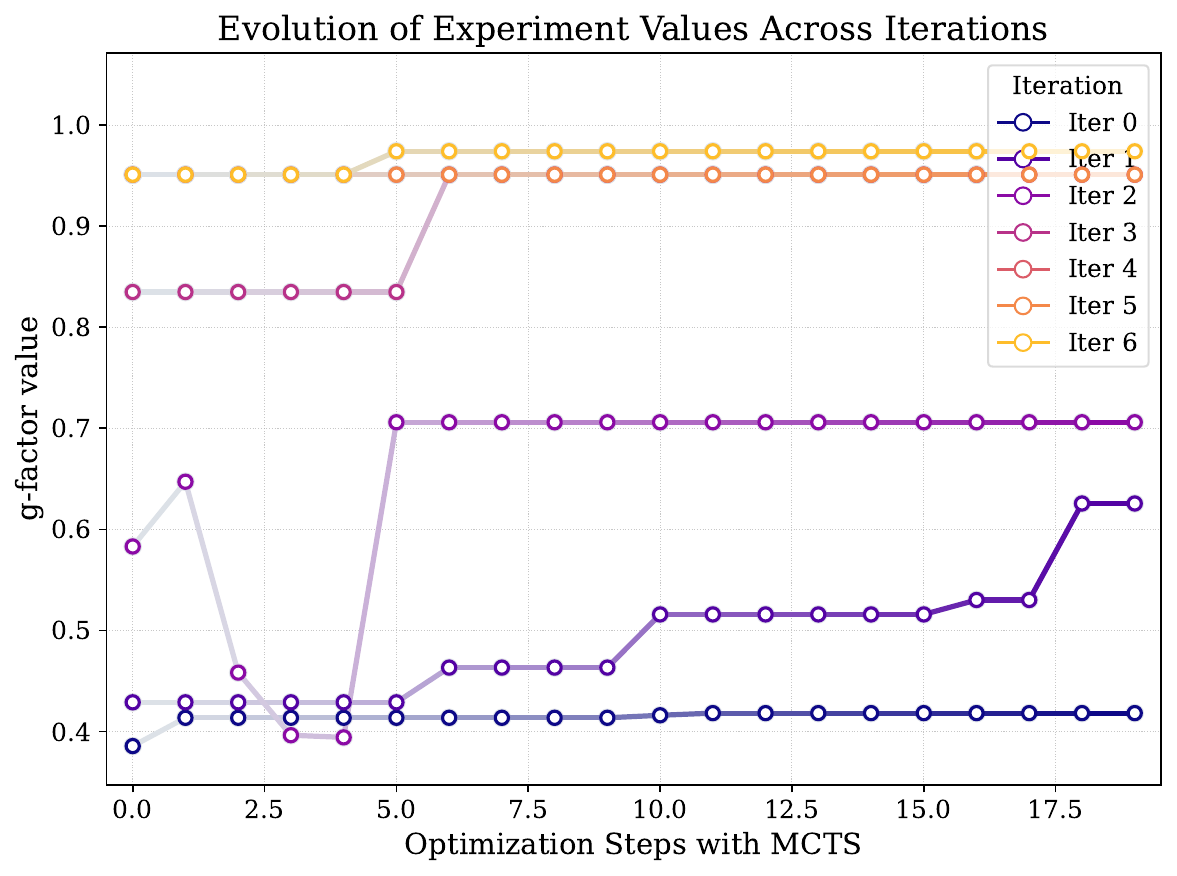}
    \caption{Evolution of Experiment Values Across Iterations by Optimizer Agent. }
    \label{fig:evolution}
    \vspace{-0.5em}
\end{wrapfigure}

We analyze the Hypothesis-related results shown in Figure~\ref{fig:case_study} and Optimizer Agent's process shown in Figure~\ref{fig:evolution}, where \texttt{PriM} incrementally adjusts the helix radius, pitch, number of turns, fiber radius, and curl parameter over multiple iterations. The initial hypothesis is guided by structural stability principles in mesogenic complexes, proposing that increasing helix radius enhances chirality due to improved molecular interactions. However, subsequent iterations reveal a complex interplay between helix geometry and chiral optical properties, necessitating further refinements to optimize g-factor performance. Figure~\ref{fig:evolution} shows the optimization progress in each iteration.

In Iteration 1, the system suggests that increasing the helix radius to $42.59$ should enhance the g-factor due to optimized structural stability and molecular interactions. However, the resulting g-factor ($0.418$) remains suboptimal. Iteration 2 refines this hypothesis, introducing the number of turns (n\_turns) as another key parameter, resulting in a g-factor improvement to $0.625$, demonstrating the interdependence between the number of turns and helical chirality.

As the iterations progress, the Hypothesis Agent and the Optimizer Agent systematically adjusts parameters based on experimental validation. In Iteration 4, the introduction of fiber radius and height parameters significantly modulates helical symmetry, leading to a g-factor of $0.95$. By Iteration 8, \texttt{PriM} arrives at an optimized configuration, achieving a maximum g-factor of $0.974$, yielding a $133\%$ improvement from the initial value. We take an example to show a detailed literature-hypothesis-experiment with our \texttt{PriM} framework:

\begin{figure}[t!]
    \centering
    \includegraphics[width=0.98\textwidth]{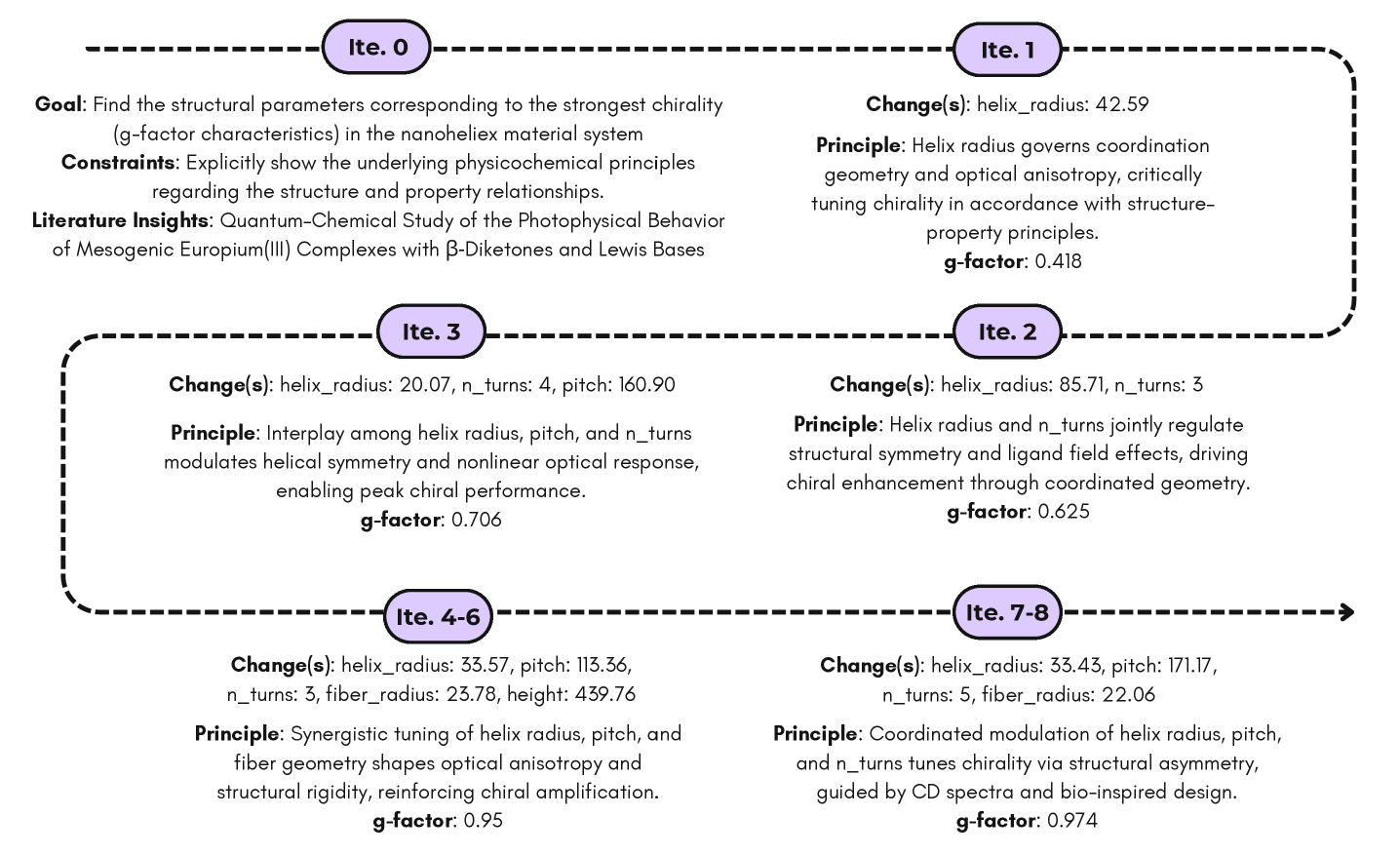}
    \caption{
        Step-by-Step Principle Evolution. Each step records the principles behind the hypothesis, the changes of parameter values, and the achieved g-factor, highlighting key improvements and showcases \texttt{PriM}'s ability to balance exploration and exploitation.
    }\label{fig:case_study}
\end{figure}

\noindent \textbf{Literature Insights.} Before the hypothesis is proposed, some insights are summarized by Literature Agent. In this example, a paper \textit{Quantum-Chemical Study of the Photophysical Behavior of Mesogenic Europium(III) Complexes with $\beta$‐Diketones and Lewis Bases} with its summarization establish that \textbf{coordination polyhedra govern optical properties} in complex materials. This provides the foundation for the hypothesis that helix radius directly influences chirality (g-factor) in nanohelices. The relationship stems from three key principles: first, chirality represents a fundamental optical property in helical nanostructures; second, coordination geometry significantly alters light-matter interactions in helical arrangements; and third, the radius parameter precisely determines the spatial organization of coordination sites.

\noindent \textbf{Hypothesis Generation.} Based on the literature review, the generated hypothesis is: \textit{By optimizing the helix radius to an initial value of 55 (within the range of 20 to 90), the nanohelices material system will exhibit the strongest chirality (g-factor characteristics), as the helix radius significantly influences the coordination polyhedra and optical properties, aligning with the physicochemical principles of structure-property relationships highlighted in the literature.} This can also interpret why the system choose helix radius as the first attempt.

\noindent \textbf{Analysis Report.} The investigation after the hypothesis, experiment and optimization demonstrates a systematic approach from theory to experimental verification. As the provided report says, \textit{the literature review identified coordination polyhedra as critical determinants of optical properties, particularly in materials with helical geometries. This understanding informed the hypothesis that helix radius optimization (predicted at 55 nm) would maximize chirality (g-factor).} Subsequent experiments revealed a more complex relationship than initially proposed, with optimal chirality occurring at 42.59 nm. This 22.6\% deviation from our hypothesis, coupled with the contradictory correlation coefficients (negative Pearson but positive Spearman/Kendall), suggests that while the fundamental principle candidates was correctly identified, the actual behavior follows a non-linear pattern that requires more sophisticated modeling beyond our initial linear prediction framework.

\noindent \textbf{System Effectiveness.} The \texttt{PriM} iterative research system has demonstrated remarkable efficacy in optimizing nanohelix materials for enhanced chirality. Through successive cycles of literature analysis, hypothesis generation, and experimental validation, the system refined its \textit{understanding} of structure-property relationships—evolving from an initial hypothesis predicting optimal helix radius at 55 nm to discovering the actual optimum at 33.43 nm with a maximum g-factor of 0.974. This progression showcases \texttt{PriM}'s ability to efficiently navigate complex parameter spaces by integrating theoretical insights from coordination chemistry with empirical data validation. Most notably, our system identified critical synergistic relationships between multiple structural parameters (pitch, helix radius, curl, and number of turns) that would have been difficult to discern through traditional research methodologies. The average 24.85\% improvement in g-factor between iterations validates \texttt{PriM}'s scientific approach and highlights its potential for accelerating materials discovery across diverse technological domains.

The complete iterative refinement process is summarized in Figure~\ref{fig:case_study}, which illustrates the evolution of hypotheses, suggested conditions from MCTS, and their corresponding optimal g-factor values, alongside the physicochemical principles underlying each refinement.

\section{Impact Statement}
To the best of our knowledge, this work represents the first exploration of principle-driven materials discovery (PMD) using language models. Our \texttt{PriM} framework demonstrates a significant advancement in automated materials discovery by integrating scientific principles into the exploration process through multi-agent collaboration. By embedding physicochemical principles into hypothesis generation and experimental validation, \texttt{PriM} not only achieves superior performance in identifying optimal material properties but also maintains interpretability throughout the discovery process. This approach bridges the gap between black-box optimization and scientific understanding, paving the way for AI-assisted discovery that remains grounded in established scientific methodology.

\section{Limitations and Future Work}
We acknowledge several limitations in our current implementation. The reliance on LLM inference introduces potential biases or hallucinated correlations between structure variables and property values. Although \texttt{PriM} includes verification mechanisms, further work is needed to quantify and mitigate these biases. Additionally, while our case study focused on nanohelix optimization, broader validation across diverse material systems would strengthen our claims about generalizability.

Future work will explore several promising directions. We plan to incorporate advanced reasoning mechanisms that refines hypothesis selection through structured causal inference, potentially improving exploration efficiency. Integration with automated laboratory platforms would enable physical validation of our approach in real-world settings. Furthermore, we aim to extend \texttt{PriM} to more complex scientific systems where mechanistic understanding is limited, such as complex catalytic reactions or biological materials, where principle-guided exploration could be particularly valuable. By bridging automated reasoning with scientific knowledge, \texttt{PriM} represents a step toward more interpretable, efficient, and scientifically grounded AI systems for materials discovery.

%
%
\section{Conclusions}
This work presents \texttt{PriM}, a principle-guided multi-agent system that advances materials discovery by embedding scientific reasoning within a structured hypothesis-validation framework. Unlike traditional black-box searching approaches, \texttt{PriM} increases interpretability and transparency, balancing exploration and exploitation. With this approach, materials with targeting property are not only discovered but also validated through suggested physicochemical principles. Our experiments about nano helix material discovery demonstrate that \texttt{PriM} accelerates discovery efficiency, yielding higher optimal property values compared to single-agent and conventional multi-agent baselines. The Hypothesis Agent plays a crucial role in refining search trajectories while avoiding unproductive exploration, as confirmed by our ablation studies. By bridging automated reasoning with scientific knowledge, \texttt{PriM} lays the groundwork for the next generation of AI-driven discovery frameworks, fostering faster, more reliable, and more interpretable scientific innovation.

\bibliography{original}
\bibliographystyle{original}

\newpage

\appendix

\section{Prompt Engineering}\label{sec:prompt_engineering}
\subsection{User Proxy Agent}
\subsubsection{Summarize Research Goal}
\begin{tcolorbox}[colback=blue!5!white,colframe=blue!75!black,title=System Prompt]
Now you are an expert in research goal refinement for scientists. You should clarify and summarize research goals to make them precise and suitable for querying scientific databases like the Semantic Scholar API.
\end{tcolorbox}
\begin{tcolorbox}[colback=gray!5!white,colframe=gray!75!black,title=User Prompt]
Here, I would like to refine a given research goal for clarity and specificity. I need you to \textbf{generate a clarified research goal} by following these requirements:
\begin{itemize}
    \item Maintain all critical scientific details and domain-specific terminology.
    \item Ensure the clarified goal is concise and uses keywords relevant to the research context.
    \item Remove extraneous or general descriptive phrases.
    \item Align the clarified goal with requirements for effective querying using scientific databases.
    \item Format the clarified goal as a concise, keyword-focused statement.
\end{itemize}
\end{tcolorbox}

\subsubsection{Summarize Research Constraints}
\begin{tcolorbox}[colback=blue!5!white,colframe=blue!75!black,title=System Prompt]
Now you are an expert in research constraint refinement for scientists. You should clarify and summarize research constraints to make them precise and suitable for querying scientific databases.
\end{tcolorbox}
\begin{tcolorbox}[colback=gray!5!white,colframe=gray!75!black,title=User Prompt]
Here, I would like to refine the constraints of a research project for clarity and specificity. I need you to \textbf{generate clarified research constraints} by following these requirements:
\begin{itemize}
    \item Identify and emphasize the key limitations and boundaries of the research project.
    \item Ensure the clarified constraints are concise and use domain-specific terminology.
    \item Remove redundant or overly general phrases that do not contribute to a specific understanding of the constraints.
    \item Align the clarified constraints with requirements for effective querying using scientific databases.
    \item Format the clarified constraints as a concise, keyword-focused statement.
\end{itemize}
\end{tcolorbox}

\subsection{Literature Agent}
\subsubsection{Get Semantic Scholar Search Keywords}
\begin{tcolorbox}[colback=blue!5!white,colframe=blue!75!black,title=System Prompt]
Now you are an expert in generating search keywords for scientific database queries. Your task is to use refined research goals and research constraints to create precise and effective search queries for the Semantic Scholar API in the required format.
\end{tcolorbox}

\begin{tcolorbox}[colback=gray!5!white,colframe=gray!75!black,title=User Prompt]
Here, I need you to generate search queries for a literature review. Please follow these requirements:
\begin{itemize}
    \item Use the provided clarified research goal and constraints to identify relevant search queries.
    \item Ensure the output is formatted as a single search string separated by commas, suitable for the Semantic Scholar API.
    \item Maintain brevity and precision, using domain-specific terms.
    \item Ensure the search terms cover both the research goal and constraints effectively.
    \item The query words you suppose should be as few as possible, as Semantic Scholar may not find enough literature with too many constraints.
\end{itemize}
\end{tcolorbox}

\subsubsection{Summarize Searched Results}
\begin{tcolorbox}[colback=blue!5!white,colframe=blue!75!black,title=System Prompt]
You are an expert in summarizing literature review results from scientific database searches. Your task is to process and summarize results retrieved from the Semantic Scholar API, focusing on the \textbf{mechanisms} by which various factors affect nanohelices materials.
\end{tcolorbox}

\begin{tcolorbox}[colback=gray!5!white,colframe=gray!75!black,title=User Prompt]
Here, I would like to summarize the search results from a literature review. The summaries should focus on the \textbf{mechanisms} and their \textbf{impact on nanohelices materials}. Please adhere to the following REQUIREMENTS:
\begin{itemize}
    \item Include the article title, authors, and publication year.
    \item Provide a 1–2 sentence summary of the article's focus on \textbf{mechanisms}, specifically how different factors or processes affect \textbf{nanohelices materials}.
    \item Use precise scientific language to ensure clarity and relevance.
    \item Avoid including unrelated details; prioritize findings directly tied to the effects on nanohelices materials.
    \item Format the summaries for easy reference and further exploration.
\end{itemize}
\end{tcolorbox}

\subsection{Hypothesis Agent}
\subsubsection{Generate Hypothesis}
\begin{tcolorbox}[colback=blue!5!white,colframe=blue!75!black,title=System Prompt]
You are an expert in materials science with a focus on helical structures and chiral properties. Your task is to generate clear, specific, and testable hypotheses for nanohelices research. Each hypothesis should be grounded in scientific principles of helix geometry, chirality, and material behavior, and it must guide the design of experiments to evaluate these properties. Incorporate insights from literature, ensure alignment with research goals and constraints, and propose parameters within the defined space for virtual experiments.
\end{tcolorbox}

\begin{tcolorbox}[colback=gray!5!white,colframe=gray!75!black,title=User Prompt]
Here, I would like to generate a clear and testable hypothesis based on the provided research information. Please follow these REQUIREMENTS:
\begin{itemize}
    \item Ensure the hypothesis aligns with the given research goal.
    \item Address all the specified research constraints.
    \item Incorporate insights or patterns identified in the provided literature review.
    \item Specifically consider the principles of helix geometry and chirality in the hypothesis.
    \item Focus on testing one parameter from the provided parameter space that is most relevant to the research goal.
    \item Include the parameter label and an initial value for the experiment, supported by literature or logical reasoning.
    \item Format the output as a single CONCISE hypothesis statement.
\end{itemize}
\end{tcolorbox}

\subsubsection{Refine Hypothesis}
\begin{tcolorbox}[colback=blue!5!white,colframe=blue!75!black,title=System Prompt]
You are an expert in refining hypotheses for nanohelices research. Your primary task is to enhance hypotheses by incorporating insights from the research report of the previous iteration of experiments, and theoretical principles related to helix structure and chirality. The refined hypothesis must be precise, TESTABLE, and explicitly address the research objectives, constraints, and experimental outcomes. Pay special attention to the interplay between helix geometry (e.g., pitch, n\_turns, helix\_radius) and material properties, such as mechanical strength, optical activity, and chirality. Where applicable, use the Circular Dichroism (CD) spectrum as a guiding factor. Propose adjustments for future experiments to validate the hypothesis and explore hidden connections among parameters.
\end{tcolorbox}

\begin{tcolorbox}[colback=gray!5!white,colframe=gray!75!black,title=User Prompt]
Based on the following research report from the previous iteration of experiments, refine the hypothesis to better align with the research goal, constraints, and experimental outcomes. The hypothesis you revised MUST be CONCISE!

The refined hypothesis must:
\begin{itemize}
    \item Be CONCISE and focused on a specific parameter from the given parameter space:
        \begin{itemize}
            \item \verb|angle: [0.123160654, 1.009814211]|
            \item \verb|curl: [0.628318531, 8.078381109]|
            \item \verb|fiber_radius: [20, 60]|
            \item \verb|height: [43.32551229, 954.9296586]|
            \item \verb|helix_radius: [20, 90]|
            \item \verb|n_turns: [3, 10]| (integer values only)
            \item \verb|pitch: [60, 200]|
            \item \verb|total_fiber_length: [303.7757835, 1127.781297]|
            \item \verb|total_length: [300, 650]|
        \end{itemize}
    \item Clearly articulate how the selected parameter influences material properties and contributes to achieving the research goal.
    \item You may suggest parameters within the defined space for virtual experiments.
    \item Apart from the experiment variables from the past iteration, you are encouraged to consider other parameters from the parameter space.
    \item Suggest specific values or adjustments for the parameter based on supporting evidence from experiments or literature.
    \item Explore potential hidden connections or interdependencies among parameters and propose hypotheses to investigate them.
    \item Format the output as a single CONCISE hypothesis statement.
\end{itemize}
\end{tcolorbox}

\subsection{Experiment Agent}
\subsubsection{Initialize Experiment}
\begin{tcolorbox}[colback=blue!5!white,colframe=blue!75!black,title=System Prompt]
Now you are an expert in designing scientific experiments. Your task is to identify the experimental variables to be tested based on a given hypothesis. The output must include the parameter names and their proposed initial values from the hypothesis.
\end{tcolorbox}

\begin{tcolorbox}[colback=gray!5!white,colframe=gray!75!black,title=User Prompt]
Here, I need you to identify the experimental variables from a hypothesis. Please follow these REQUIREMENTS:
\begin{itemize}
    \item Extract the specific parameters to be tested from the hypothesis, and the initial values of these parameters MUST be NUMERICAL values.
    \item ONLY output the parameter names and their initial values in the format: \verb|{'variables': ['var1', 'var2'], 'values': [val1, val2]}|, do not include anything else.
    \item The variable and corresponding parameter you are suggesting MUST lie in the pre-defined parameter space:
        \begin{itemize}
            \item \verb|angle: [0.123160654, 1.009814211]|
            \item \verb|curl: [0.628318531, 8.078381109]|
            \item \verb|fiber_radius: [20, 60]|
            \item \verb|height: [43.32551229, 954.9296586]|
            \item \verb|helix_radius: [20, 90]|
            \item \verb|n_turns: [3, 10]| (integer values only)
            \item \verb|pitch: [60, 200]|
            \item \verb|total_fiber_length: [303.7757835, 1127.781297]|
            \item \verb|total_length: [300, 650]|
        \end{itemize}
\end{itemize}
\end{tcolorbox}

\subsection{Analysis Agent}
\subsubsection{Generate Research Report}
\begin{tcolorbox}[colback=blue!5!white,colframe=blue!75!black,title=System Prompt]
    You are a research report writer specializing in materials science experiments.
\end{tcolorbox}

\begin{tcolorbox}[colback=gray!5!white,colframe=gray!75!black,title=User Prompt]
You are now requested to compile a comprehensive research report based on our research settings, experiment results, and analysis.

\textbf{Research Context}:
\begin{itemize}
    \item \textbf{Research Goal}: \verb|research_goal|
    \item \textbf{Constraints}: \verb|research_constraints|
    \item \textbf{Literature Review Summary}: \verb|literature_insights|
    \item \textbf{Hypothesis}: \verb|hypothesis|
\end{itemize}

\textbf{Data Analysis}:
The analysis results obtained from the data analysis tools are attached below: \verb|data_analysis_results|.

\textbf{Requirements for the Report}:
    \begin{itemize}
        \item Provide a \textbf{concise summary} of the experimental results.
        \item Highlight important \textbf{insights} from the data and analysis.
        \item Include \textbf{tables} summarizing experimental setups, key parameters, and results.
        \item Suggest \textbf{next steps} for the research based on the current findings.
    \end{itemize}

The report should be saved as a structured markdown file. AND THE REPORT MUST BE CONCISE!

Make sure the report is well-structured, easy to read, and conveys the necessary details for further analysis and replication.
\end{tcolorbox}

\section{Baseline Implementations}\label{sec:baseline}

In this section, we detail the implementation of the baselines used for comparison, namely the \textbf{Vanilla Agent} and the \textbf{Vanilla MAS}.

\subsection{Vanilla Agent}
We define the Vanilla Agent baseline as
\[
\mathcal{F}_{\text{SLLM}} = \{E, \mathcal{O}\},
\]
where:
\begin{itemize}
    \item $E$ is the \textbf{Experiment Agent} that executes experiments based on the predefined research goal.
    \item $\mathcal{O}$ is the \textbf{Optimizer Agent} that refines the experimental parameter space via Monte Carlo Tree Search (MCTS)-based optimization.
\end{itemize}
In this baseline, the hypothesis generation phase is omitted, and the resultant g-factor value $\mathcal{D}_t$ at iteration $t$ is produced directly as:
\[
\mathcal{D}_t = E(\mathcal{X}_t),
\]
where $\mathcal{X}_t$ is the parameter space at the $t$-th iteration. And $\mathcal{X}_1$ is defined according to the pre-defined research goal and constraints.

The iterative MCTS-based parameter update is then given by:
\[
\mathcal{X}_{t+1} = \mathcal{O}\left(\mathcal{X}_t, \mathcal{D}_t\right),
\]
where the Optimizer Agent continuously adjusts $\mathcal{X}_t$ based on observed outcomes.

\subsection{Vanilla MAS}
The Vanilla MAS baseline is defined as
\[
\mathcal{F}_{\text{MAS}} = \{\mathcal{P}, \mathcal{U}, L, H, E, A, \mathcal{O}\},
\]
where:
\begin{itemize}
    \item $\mathcal{P}$ is the \textbf{Planner} agent managing overall workflow and agent coordination.
    \item $\mathcal{U}$ is the \textbf{User Proxy Agent} that specifies the research goals and constraints.
    \item $L$ is the \textbf{Literature Agent} which retrieves and summarizes relevant literature.
    \item $H$ is the \textbf{Hypothesis Agent} that, in the full framework, generates testable propositions; however, in this ablated baseline, the operational role of $H$ is removed.
    \item $E$ is the \textbf{Experiment Agent} that executes experiments.
    \item $A$ is the \textbf{Analysis Agent} that processes experimental data and generates analysis reports.
    \item $\mathcal{O}$ is the \textbf{Optimizer Agent} that optimizes the parameter space.
\end{itemize}
In the ablated Vanilla MAS, by bypassing the hypothesis generation, the workflow is defined by:
\[
\mathcal{S}_{t+1} = \mathcal{P}\big(\mathcal{S}_t, \mathcal{R}_t\big),
\]
with operations occurring as:
\begin{enumerate}
    \item The User Proxy Agent $\mathcal{U}$ sets the research goal and constraints.
    \item The Literature Agent $L$ extracts insights from scientific literature.
    \item The Hypothesis Agent $H$ is bypassed, and the system proceeds directly to the Experiment Agent $E$.
    \item The Optimizer Agent $\mathcal{O}$ refines the parameter space $\mathcal{X}_t$ based on experimental feedback.
    \item The Analysis Agent $A$ evaluates experimental outcomes, generating report $\mathcal{R}_t$.
\end{enumerate}

\section{Experiment Data Analysis}\label{sec:experiment_data_analysis}
To quantitatively analyze the experimental data and extract meaningful insights, we employ a suite of statistical and computational tools. The \textbf{Analysis Agent} utilizes these tools to process experimental records, determine statistical distributions, compute correlation metrics, fit polynomial models, and identify optimal experimental conditions. The following subsections provide a detailed mathematical description of these methods.
    
\subsection{Data Distribution Analysis}
To access the variability and normalization of experimental parameters, we employ \textbf{standardization} techniques. Given a dataset of experimental variables $X = (X_1, \ldots, X_n)$ and the corresponding experimental results $Y = (y_1, \ldots, y_n)$, the standardized value $X_i^\prime$ for each parameter is computed as:
    \[X_i^\prime = \frac{X_i - \mu_X}{\sigma_X}, \quad Y^\prime = \frac{Y - \mu_Y}{\sigma_Y},\]
    where $\mu_X$ and $\sigma_X$ are the mean and standard deviation of $X$, and $\mu_Y$, $\sigma_Y$ are those of $Y$. This transformation ensures that each parameter follows a standard normal distribution, facilitating fair comparisons across different experimental variables.
    
\subsection{Correlation Analysis}
    To evaluate the dependence between experimental parameters and outcomes, we compute three correlation coefficients:

\subsubsection{Pearson Correlation Coefficient}
    The Pearson correlation coefficient $r_p$ measures the linear dependence between a variable $X$ and a response $Y$:
    \[r_p = \frac{\sum_{i=1}^n (X_i - \bar{X})(Y_i - \bar{Y})}{\sqrt {\sum_{i=1}^n (X_i \bar X)^2} \sqrt {\sum_{i=1}^n (Y_i - \bar Y)^2}},\]
    where $\bar X$ and $\bar Y$ are the mean values of $X$ and $Y$, respectively.

\subsubsection{Spearman Rank Correlation}
    The Spearman correlation coefficient $r_s$ evaluates the \textbf{monotonic relationship} between two variables by ranking values before computing the Pearson correlation:
    \[r_s = 1 - \frac{6\sum d_i^2}{n(n^2 - 1)},\]
    where $d_i$ is the difference between the ranks of corresponding $X$ and $Y$ values.
    
\subsubsection{Kendall's Tau}
    Kendall's Tau $\tau_K$ assesses the strength of ordinal association:
    \[\tau_K = \frac{C - D}{C + D}, \] where $C$ and $D$ are the number of concordant and discordant pairs.

    A pair $(x_i, y_i)$ and $(x_j, y_j)$ is said to be concordant if $(x_i - x_j)(y_i - y_j) > 0$, is said to be discordant if $(x_i - x_j)(y_i - y_j) < 0$.

\subsection{Critical Value Identification}
    To determine the optimal experimental conditions, we extract the maximum experimental result $g = \max_i Y_i$, the corresponding optimal parameter is given by \[X^\star = \{X_{i^\star} \mid i^\star = \argmax_i Y_i\},\] which identifies the parameter values associated with highest observed response.

\subsection{Polynomial Curve Fitting}
    To model the nonlinear relationship between experimental variables and the g-factor results, we perform polynomial regression. Given an independent variable $X$ and the response $Y$, we fit a polynomial function of degree $d$:
    \[Y \approx f(X) = c_d X^d + c_{d-1} X^{d-1} + \ldots + c_1 X + c_0, \] where the coefficients $c_0, c_1, \ldots, c_d$ are obtained using least squares fitting:\[\min_{c_0, c_1, \ldots, c_d} \sum_{i=1}^d (Y_i - f(X_i))^2.\]

\section{Virtual Lab}\label{sec:virtual_lab}
The Virtual Lab is a computational environment designed to evaluate material properties based on structural parameters. Preparing the virtual lab consists of data preprocessing, model training, and inference components to predict the g-factor of nanomaterials.

We develop a http server based on flask app to host an API to support the numerical experiments. All requests of getting property value is through the local http server.

\end{document}